\documentclass[accepted]{uai2025}                       
\usepackage[american]{babel}

\usepackage{natbib} 
\usepackage{mathtools} 
\usepackage{bm}
\usepackage{makecell}
\usepackage{multirow}
\usepackage{booktabs} 
\usepackage{tikz} 

\title{Uncertainty-Aware Decomposed Hybrid Networks}

\author[1,2]{\href{mailto:<sina.ditzel@flandersmake.be>}{Sina Ditzel\thanks{
This work was conducted at Goethe university and 
the time for subsequent writing was supported by Flanders Make.
}}{}}
\author[1]{\href{mailto:<Jaziri@em.uni-frankfurt.de>}{Achref Jaziri}{}}
\author[1,3]{\href{mailto:<pliushch@psych.uni-frankurt.de>}{Iuliia Pliushch}{}}
\author[1]{\href{mailto:<vramesh@em.uni-frankfurt.de>}{Visvanathan Ramesh}{}}

\affil[1]{%
    Department of Computer Science and Mathematics\\
    Goethe University\\
    60323 Frankfurt am Main, Germany\\
}
\affil[2]{%
    Flanders Make,
    Oude Diestersebaan 133,
    3920 Lommel,
    Belgium
}  
\affil[3]{ Department of Educational Psychology, Goethe University, 60629 Frankfurt am Main, Germany
}
  
\begin{document}
\maketitle

\begin{abstract}

    The robustness of image recognition algorithms remains a critical challenge, as current models often depend on large quantities of labeled data. In this paper, we propose a hybrid approach that combines the adaptability of neural networks with the interpretability, transparency, and robustness of domain-specific quasi-invariant operators. Our method decomposes the recognition into multiple task-specific operators that focus on different characteristics, supported by a novel confidence measurement tailored to these operators. This measurement enables the network to prioritize reliable features and accounts for noise. We argue that our design enhances transparency and robustness, leading to improved performance, particularly in low-data regimes. Experimental results in traffic sign detection highlight the effectiveness of the proposed method, especially in semi-supervised and unsupervised scenarios, underscoring its potential for data-constrained applications.
\end{abstract}

\section{Introduction}\label{sec:intro}
\begin{figure*}
	\centering
	\includegraphics[width=\linewidth]{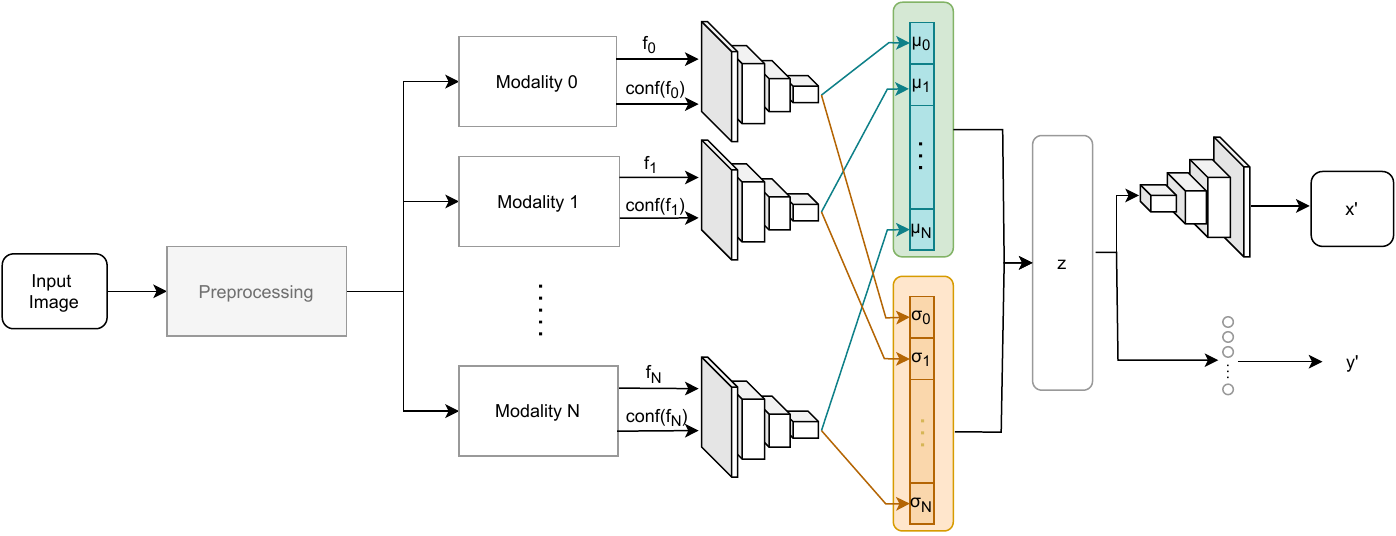}
	\caption{
    The proposed hybrid architecture decomposes the input image $x$ using task-specific quasi-invariances $T_{i}(x)$, with computed uncertainties $conf(T_{i}(x))$. We experimented with different neural architectures that propagate the confidences through initial layers via normalized convolution. The here visualized VAE based architecture encodes the transformed signal into a joint latent space z (with mean $\mu$ and standard deviation $\sigma$), which is then used for tasks, such as classification y.}

	\label{fig:outline_pipeline}
\end{figure*}
Deep neural networks have proven to be highly effective in scenarios where training data sufficiently covers the underlying distribution of the deployment domain. 
Despite their success, DNNs function as black-box models, limiting interpretability. This opacity can lead to unintended biases, such as reliance on spurious correlations in object placement \citep{pinto08} or biases related to skin tone and gender\citep{Balakrishnan2020}, that are difficult to reveal. Further, this highlights that these models often struggle to generalize to unseen conditions. Real-world applications, however, involve diverse variations, such as changes in pose, lighting, weather, and sensor quality, which are cumbersome to cover in the dataset.
Addressing these challenges requires improving model transparency and robustness. 
In this work, we propose an approach towards more transparent systems by integrating quasi-invariants, supported explicitly by a confidence measure with neural networks, which we believe can enhance robustness and interpretability.

DNNs implicitly learn transformations for classification without explicitly enforcing invariances \citep{bengio2009learning, krizhevsky2012imagenet}. While effective, this lack of explicit control reduces interpretability and increases vulnerability to adversarial attacks \citep{barredoarrieta2020}. To mitigate these issues, research has explored decomposable designs \citep{Lipton2016} and targeted specific invariance, typically needed in computer vision applications. 
Early model-based approaches leveraged quasi-invariant transformations to ensure robustness \citep{binford1993quasi, chin1986model}. Unlike purely data-driven models, quasi-invariant operators offer inherent degree of transparency.
The integration of these operators allows for selective suppression of nuisance variables while preserving task-relevant features \citep{baslamisli2021}. Hybrid designs, which embeds predefined priors as fixed initializations in the first layers of neural networks, have shown improvements in robustness \citep{dapello2020simulating, scatteringhybrid}. 
Recent works \citep{luan2018, jaziri2023representation} demonstrate that incorporating  quasi-invariant filters, such as Gabor filters and decomposing the input into various parallel streams enhances neural network robustness and representation learning. This also aligns with neuroscience principles, as decomposable vision systems resemble brain-inspired architectures \citep{von2014vision, HotaRamesh2013}. However, integrating quasi-invariances necessitates complementary uncertainty measures to further refine the decision-making process. 

Basic neural networks do not provide confidence estimates and often exhibit over- or under-confidence, leading to miscalibrated predictions. Reliable uncertainty estimates are crucial, particularly in high-risk applications where incorrect predictions can have serious consequences. Such estimates allow uncertain predictions to be ignored or referred to human experts \citep{abdar2021review}. Uncertainty estimation is also essential in domains with highly heterogeneous data sources and limited labeled data.


In this work, we illustrate the design of a neural network architecture that leverages inductive biases in the form of known invariance requirements for the task, sensor noise models, decompositions of input data to sub-modalities such as color, texture. We propose a novel confidence measure. Uncertainty modeling and statistical hypotheses tests allow for estimation of posterior probabilities of given sub-modality features that serve as weights in the encoders, producing latent representations in a Variational encoder-decoder architecture formulation (see Figure ~\ref{fig:outline_pipeline}). 
The task-specific operators are based on invariance properties \citep{lbp_survey, gevers_color_99}, while the confidence measure builds on prior work in sensor noise modeling \citep{forstner2000image, noise_immerkaer} and uncertainty propagation through chosen operators \citep{greiffenhagen_2001}. We leverage normalized convolutions \citep{NCNN} to integrate the confidence measure as weight. The design may be seen as a fusion of human devised statistical models and systems analysis along with modern deep learning architectures to enhance transparency, robustness, and explainability of results. 

%


\section{Related Work}\label{sec:related_work}

\textbf{DNNs with Invariant Operators:} Several studies have explored the integration of well-established operators into data-driven neural networks to enhance transparency and robustness \citep{li2020wavelet}. For instance, the SIFT detector, which provides quasi-invariance to scale, orientation, and illumination, has been incorporated into neural architectures to improve feature extraction \citep{perronnin2015fisher}. Similarly, adaptations of the Local Binary Pattern operator have been employed to design convolutional layers with reduced parameter complexity, effectively capturing texture features \citep{juefei2017local}. Further, \citet{dapello2020simulating} has shown that simulating the functionality of the primary visual cortex in the early layers of CNNs enhances robustness against input perturbations and adversarial attacks. 
These methods collectively demonstrate the advantages of integrating structured, interpretable transformations with deep learning models to achieve better robustness and trust. Building on these works, we propose a general framework for decomposable designs incorporating uncertainty estimation of the chosen operators. Our uncertainty aware decomposable design further enhances model reliability and adaptability in diverse application contexts.

\textbf{Uncertainty Estimation in Neural Networks:}
A key distinction exists between aleatoric uncertainty, which arises from inherent data variability, and epistemic uncertainty, which stems from model limitations \citep{hacking2006emergence}. Aleatoric uncertainty is irreducible, whereas epistemic uncertainty can be minimized with additional training data. 
Probabilistic methods estimate uncertainty by generating an ensemble of models and quantifying uncertainty through variance or entropy over their predictions \citep{depeweg2018decomposition}. Deep ensembles \citep{lakshminarayanan2017simple} explicitly train multiple models to capture epistemic uncertainty but are computationally expensive. To mitigate this overhead, alternative approaches approximate deep ensembles without requiring multiple model instances. Recent works on mode connectivity \citep{garipov2018loss} suggest that ensembles can be optimized along low-loss paths in the weight space, reducing the computational overhead during ensemble training.
Bayesian neural networks (BNNs) \citep{mackay1992practical} employ variational inference \citep{zhang2018advances} to model posterior weight distributions, while Monte-Carlo (MC) dropout \citep{gal2016dropout} introduces stochasticity through dropout layers to induce variability in predictions. However, these methods come with trade-offs: BNNs incur high runtime costs due to their parameter-heavy nature or can degrade prediction quality.
Another line of research explores distribution-free uncertainty estimation using conformal prediction \citep{balasubramanian2014conformal} and quantile regression \citep{angelopoulos2022image} to estimate aleatoric uncertainty bounds.
In our approach, aleatoric uncertainty is linked to camera noise, while epistemic uncertainty is managed adaptively within a Bayesian framework by dynamically updating priors. 

\textbf{Noise Awareness in Model-Based Design:}
In traditional computer vision, many studies focus on estimating noise to minimize its effects. \citet{boncelet2009image} reviews noise models and recommendations for mitigation. \citep{tsin2001statistical} showed that careful modeling and calibration of the camera response function and its noise sources achieve robust illumination invariant change detection. In other works, sensor noise is often approximated using additive Gaussian Noise, which can be estimated from a single image \citep{forstner2000image}, with \citet{noise_immerkaer} offering a fast algorithm for this.  
\citet{gevers2004robust} and \citet{, greiffenhagen_2001} show that propagating the estimated Gaussian noise through color invariant operators improves robustness for recognition systems. Our work is complementary to these approaches as we focus on modeling uncertainty of the chosen operators  integrated into data-driven models. We demonstrate how this design paradigm can improve performance, particularly in applications where prior knowledge can inform model structure and uncertainty estimation.

\section{Decomposed Network with Confidence Estimation}\label{sec:detailed-system}
\begin{figure*}[t]
	\centering 
	\includegraphics[width=\linewidth]{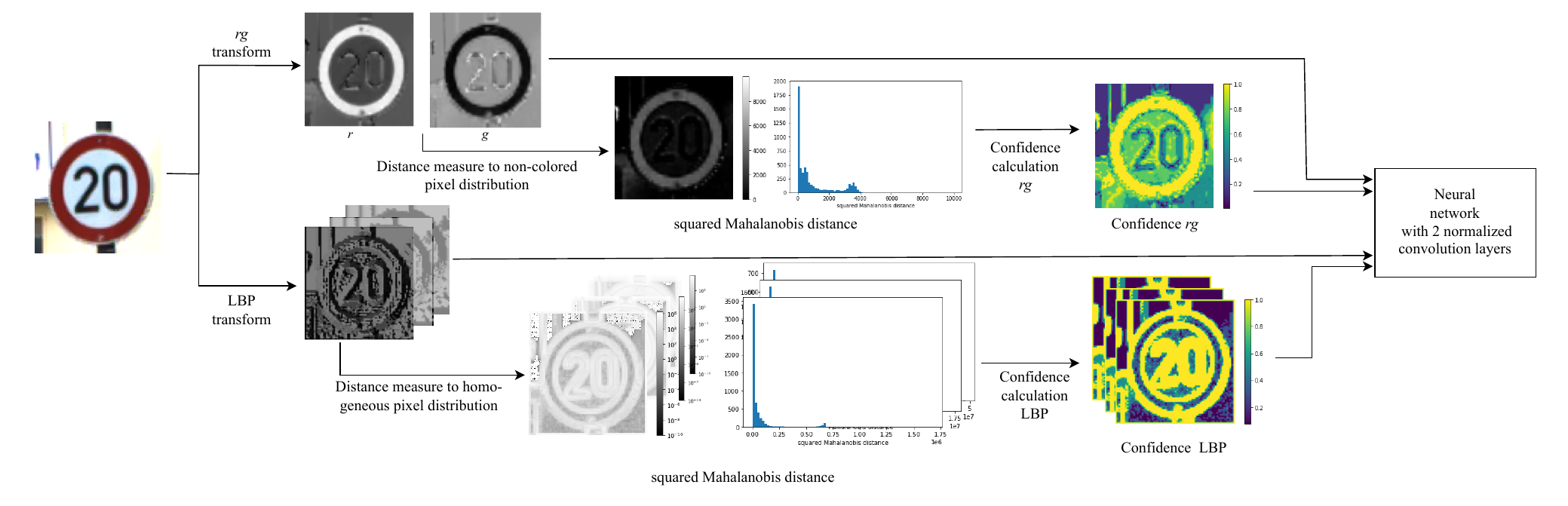}
	\caption{Example of our method applied to the GTSRB dataset using the \textit{rg} and LBP operators. The figure shows key intermediate steps, including operator outputs and confidence computation via Mahalanobis distance, calculated per pixel and for visualization also aggregated into a histogram. Both the transformed image and its confidence are propagated through the neural network for a downstream task. Confidence is shown in blue (low) to yellow (high).}
	\label{fig:rg-LBP-Pipeline}
\end{figure*}
In this section, we introduce a decomposed uncertainty-aware hybrid design illustrated in Figure \ref{fig:outline_pipeline}.
For each application, different features are relevant. and choosing appropriate (quasi-) invariant transformations is crucial. To stabilize the behavior of these features, we introduce a new confidence measuring approach based on the propagation of input noise. 

In the following chapters, we use capital letters (e.g., $X$) to represent random variables, and small letters (e.g., $x$) to denote instances and parameters. The hat symbol ($\hat{X}$ and $\hat{x}$) is used to indicate and observed variable or instance (With noise), and bold letters are used for matrices and vectors.
However, we deviate from this convention for $RGB$, $rg$, and $I$ to avoid ambiguity: Image RGB values are represented using capital letters, normalized \textit{rg} values use small letters, and intensity $I$ is also capitalized to differentiate it from indexes $i$. Also we use $N$ to give the number of samples and $\eta$ is a random variable denoting noise.

\subsection{Confidence of Operators} 
To assess the confidence and reliability of image-derived information extracted by an operator, we employ a Bayesian statistical approach that integrates prior knowledge with observed data analysis. Our method incorporates noise modeling, covariance propagation, Mahalanobis distance calculation, and Bayesian likelihood estimation to formulate a novel confidence measure. The approach is based on hypothesis testing, where transformed data from quasi-invariant operators is evaluated against a predefined null hypothesis ($H_0$), representing scenarios where information is deemed less stable or relevant. For instance, in texture and structure based analysis, homogeneous regions may be classified as less significant.

The Bayesian framework provides a probabilistic basis for assessing how well the transformed data fits the model of the null hypothesis, by calculating the confidence $\textit{Conf}$ for each value (pixel) as the posterior probability of $H_{0}$, resulting in a confidence map (Figure~\ref{fig:rg-LBP-Pipeline}).
First we compute the Mahalanobis distance $d$, which measures the deviation from the null hypothesis, and then calculate the confidence $\textit{Conf}$ for each transformed value based on the observed squared Mahalanobis distance $\hat{d}^2$:
\begin{equation}
	\begin{aligned}
		&\textit{Conf} = P(\neg H_{0}|\hat{d}^2) = \\
		&= \frac{P(\hat{d}^2|\neg H_{0})P(\neg H_{0})} {P(\hat{d}^2|H_{0}) P(H_{0})+P(\hat{d}^2|\neg H_{0})P(\neg H_{0})}.
		\label{eq:conf}
	\end{aligned}
\end{equation}
In cases where prior knowledge is available, this information can be incorporated into the calculation through the prior probabilities $P(H_{0})$ and $P(\neg H_{0})$. However, if no prior knowledge is available, the formula can be simplified to the likelihood ratio. 
\subsubsection{Noise Model and Propagation} \label{sec:noise}
To calculate the Mahalanobis distance, we need to know the distribution of the data by
first modeling the sensor noise and then propagating its covariance through the data transformation process.
We adopt a Gaussian noise model for image values, commonly used in image processing to approximate sensor noise due to its simplicity and mathematical properties. Thus we make the assumption, that the noise $\eta$ affecting the pixel values in a given channel ($R$, $G$, $B$, or intensity $I$) follows a Gaussian distribution with mean zero and a channel-specific standard deviation $\sigma_C$. For each pixel, the noise is drawn independently from this distribution. Thus, for a true pixel value $C$, the observed value $\hat{C}$ is modeled as:
\begin{align}
    \hat{C} = C + \eta_C, \quad \eta_C \sim \mathcal{N}(0, \sigma_C^2)
\end{align}
%
We assume the noise distribution can be approximated by deviations in homogeneous regions, using the approach of \citet{noise_immerkaer_extended}. 
We then propagate the noise through the operator with linear covariance propagation to estimate the uncertainty of the transformed output, following the approach of \citet{greiffenhagen_2001}. 
Based on the pre-defined mean $\bm{\mu}$ of the null hypothesis $H_0$, where the transform is unstable, and using the derived covariance propagation to calculate the covariance matrix $\bm{\Sigma}$, we then calculate the squared Mahalanobis distance $d^2$ for the transformed values $\mathbf{\hat{t}}$:
\begin{equation}
	\begin{aligned}
		d^2(\mathbf{\hat{t}}, (\bm{\mu}, \bm{\Sigma})) = (\mathbf{\hat{t}} - \bm{\mu})^\top \bm{\Sigma}^{-1} (\mathbf{\hat{t}} - \bm{\mu}).
	\end{aligned}	
        \label{eq:Mahalanobis}
\end{equation}

\subsubsection{Likelihood based on Mahalanobis Distance}
\label{section:likelihood}

We compute the likelihoods $P(\hat{d}^2|\neg H_{0})$ and $P(\hat{d}^2|H_{0})$ using the squared Mahalanobis distance $d^2$. The probability density functions (PDFs), describing the probability of observing specific values of \( d^2 \) under the corresponding hypothesis, are approximated by uniform mixtures of chi-squared distributions and remain consistent across all pixels. The relative likelihood follows from the pixel-specific $d^2$ value.

The squared Mahalanobis distance $d^2(X, (\bm{\mu}, \bm{\Sigma}))$ of a normal distributed random variable $X\sim N(\bm{\mu}, \bm{\Sigma})$ follows a $\chi^2$-distribution with degrees of freedom equal to the number of dimensions (denoted as $k$) \citep[Result~4.7]{JohnsonWichern07}. Under the defined null hypothesis $H_{0}$, this distribution holds. If X deviates from $N(\bm{\mu}, \bm{\Sigma})$ the distribution becomes noncentral $\chi^2$, with a noncentrality parameter $\lambda$ reflecting the deviation. For $P(\hat{d}^2|\neg H_{0})$ and $P(\hat{d}^2|H_{0})$ follows: 
\begin{itemize}
\item Under $H_{0}$: $d^2$ is $\chi^2$-distributed if the Gaussian model holds exactly. In practice, due to noise and approximations, it will follows a mixture of noncentral distributions with small noncentrality.
\item Under $\neg H_{0}$:  $d^2$ is a mixture of noncentral $\chi^2$ distributions, with a higher noncentrality parameter. 
\end{itemize}
In Figure~\ref{fig:rg-LBP-Pipeline}, the histogram shows the distribution of the squared Mahalanobis distance for different operators on an example image, where the two peaks correspond to the models for \( H_0 \) and \( \neg H_0 \). The peak at 0 aligns with \( H_0 \), while the peak at larger values corresponds to \( \neg H_0 \).\\
Both PDFs, $f_{H_0}(x)$ and $f_{\neg H_0}(x)$, are approximated using a uniform mixture of noncentral-$\chi^2$-distribution with noncentrality parameters $\lambda$ ranging from $\lambda_1$ to $\lambda_2$:
\begin{align}
	f_H(x, \lambda_1,\lambda_2) = \int_{\lambda_1}^{\lambda_2} \frac{1}{\lambda_2-\lambda_1} f_\chi(x,k,\lambda) \text{ d}\lambda.
	\label{f_not_H0}.
\end{align}

For \( H_0 \), this range is chosen to reflect minimal deviation from the central distribution, while for \( \neg H_0 \), the range is set to larger values of \( \lambda \), capturing the increased deviation under the alternative hypothesis. The exact choice is challenging, thus we compute multiple lambda values and merge the resulting confidence maps. See more details in Appendix~\ref{app:lambda}.\\
%
%
%
%
For ease of computation we approximated the uniform mixture  of noncentral $\chi^2$ distributions by a Gaussian distribution based on the first two moments:
\begin{align}
\mu_{H}&=k+\frac{\lambda_1+\lambda_2}{2}\\
\sigma^2_{H}&=2k+2(\lambda_1+\lambda_2)+\frac{1}{12}(\lambda_1^2+2\lambda_1\lambda_2+\lambda_2^2).
\end{align}
The proof can be found in Appendix~\ref{sec:deviation_moments}.
\subsection{Using Confidence in Neural Encoder}

As highlighted in Figure \ref{fig:outline_pipeline}, each operator is paired with an encoder. The encoders outputs are concatenated into a joint latent space, representing the combined information from all operators and serving as input for the downstream tasks.
The output of the  quasi-invariant transformations along with its confidence map are processed with normalized convolutions \citep{normalizedConv} in the initial layers of each stream. 
Feature values are weighted by their confidence scores, reducing the influence of uncertain pixels. 
This ensures that uncertain regions in the input, have a minimal impact on the output, thereby enhancing the robustness of the learned representations.
As by \citet{NCNN}, we combine normalized convolutional layers with confidence pooling. During subsampling, max pooling is applied on confidence values, using the corresponding indices to pool feature maps.

\section{Adoption and Specification for  Traffic Sign Recognition}
\label{section:gtsrb-pipeline}
\label{sec:rg+LBP}

We demonstrate and evaluate our approach on the German Traffic Sign Recognition Benchmark (GTSRB) dataset \citep{GTSRB}. 
Traffic signs exhibit variability in appearance due to changes in illumination, weather conditions, and viewpoints, making them useful for evaluating model invariances and assessing robustness in recognition tasks.\\
Distinctive design features such as shape, color, and symbols (including signs and numbers) make traffic signs highly recognizable. To leverage these properties, we test our methodology using two operators that come with illumination variance: the \textit{rg} color transform and Local Binary Patterns (LBP).
The rg transform normalizes color features to reduce illumination effects while retaining critical color information essential for distinguishing traffic signs. LBP captures texture details, encoding crucial shapes and symbols. 
For each operator, we define a null hypothesis to account for potential feature instability, characterized by its mean \( \mu \) and covariance matrix \( \Sigma \), which are used to calculate the Mahalanobis distance, based on Equation~\ref{eq:Mahalanobis}. This distance is then employed to compute a confidence map, based on Equation~\ref{eq:conf}. Below, we provide explanations for each operator. Additional details on the implementations discussed are provided in Appendix~\ref{app:implementation}.

\subsection{Color-Processing with the rg-Transform}
\label{sec:color-operator-confidence}
The normalized \textit{rg} transform is defined as:
\begin{align}
	\textit{r}&=\frac{R}{R+G+B}, & \textit{g}=\frac{G}{R+G+B}.
\end{align}
If the illumination is neutral i.e. white light, this model is know to be invariant to shadow, light intensity and light direction \citep{gevers_color_99}.\\
Because neutral illumination can not be assumed by default, a correction of the light temperature based on the von Kries model is applied, transforming the recorded color channels to a canonical illuminant through a linear scaling operation. We estimated the illumination using that traffic sign images contains uncolored pixels (i.e., white areas in traffic signs). The deviation from gray, caused by the illuminant’s color temperature, is used to approximate the illumination in normalized rg space. 
%
To estimate how the noise changes throughout the \textit{rg} transform, Greiffenhagen et al.\citep{greiffenhagen_2001} have shown that it can be characterized by the following covariance matrix
\begin{equation}
    \begin{aligned}
    	&\mathbf{\hat{\Sigma}}_{\hat{r},\hat{g}}\approx\\ 
            &\approx\frac{\sigma_I^2}{S^2}\left(\begin{array}{cc}
    		\frac{\sigma^2_R}{\sigma_I^2}(1-\frac{2R}{S})+3\frac{R^2}{S^2}&
    		-\frac{\sigma^2_G R+\sigma_R^2G}{\sigma_I^2S}+3\frac{RG}{S^2}\\
    		-\frac{\sigma^2_G R+\sigma_R^2G}{\sigma_I^2S}+3\frac{RG}{S^2}&
    		\frac{\sigma^2_G}{\sigma_I^2}(1-\frac{2G}{S})+3\frac{G^2}{S^2}\\
    	\end{array}\right)
    	\label{eq:cov_prop_rg}
    \end{aligned}    
\end{equation}
with $S=R+G+B$, $\mathbf{\sigma}_I^2=\frac{\sigma_S^2}{3}=\frac{\sigma_R^2+\sigma_G^2+\sigma_B^2}{3}$.

\paragraph{Color Confidence Calculation:} 
Around $r=g=\left(\frac{1}{3}\right)$), where the RGB values are nearly equal, the \textit{rg} transform becomes very sensitive to small changes in the RGB values. This sensitivity means that even minor variations or noise in the original RGB values can cause large changes in the $r$ and $g$ values.
Additionally, this region corresponds to colors close to white or gray, where the color information is less distinct, reducing the transform's ability to extract meaningful features.\\
Therefore, as confidence measure, we calculate the posterior probability of each pixel not to be close to the point $r=g=\left(\frac{1}{3}\right)$.
This is done by testing against the null hypothesis $H_0$:
\begin{equation}
	\begin{aligned}
		H_{0_{\textit{rg}}}: & \text{ The pixel is not colored} \\&i.e \left(\begin{array}{c}r\\g \end{array}\right) = \left(\begin{array}{c} 1/3\\1/3 \end{array}\right).
	\end{aligned}
	\label{eq:H0rg}
\end{equation}
Using the derived covariance matrix $\Sigma_{\hat{r},\hat{g}}$ for observed \textit{rg} values with noise $(\hat{r},\hat{g})^T$ we estimate the distribution of non-colored pixels as:
\begin{equation}
	\begin{aligned}
		H_{0_{\textit{rg}}}:\left(\begin{array}{c} \hat{r}\\\hat{g} \end{array}\right) \sim N\left(\left(\begin{array}{c} 1/3\\1/3 \end{array}\right),\mathbf{\Sigma}_{\frac{1}{3},\frac{1}{3}} \right).
	\end{aligned}
\end{equation}
Where $\mathbf{\Sigma}_{\frac{1}{3},\frac{1}{3}}$ is approximated by the Equation~\ref{eq:cov_prop_rg}, where $r= \frac{R}{S}=\frac{1}{3}$, and $g= \frac{G}{S}=\frac{1}{3}$ was used.
%
\subsection{Texture-Processing with LBP}
\label{sec:lbp-operator-confidence}
The LBP transform is applied to pixel intensity $I =\frac{R+G+B}{3}$. For each pixel value $I$ a set of  neighbors $I_{n_{1}},...,I_{n_{N}}$ is determined based on radius $r$ and number of points $p$. Each  neighbor is compared to the center pixel $I$ in a predefined order along the circle of  neighbors. A binary number $\mathbf{B}$ is generated, where 0 denotes an intensity lower than the center pixel ($ I_{n_{i}}-I<0 \Rightarrow B_i = 0$) and 1 otherwise. Converting this binary number to decimal gives the center pixel's LBP value.
For the GTSRB dataset the LBP Operator was used with 3 different sizes. The original parametrization with $r_1 = 1$, $p_1 = 8$ was complemented by $r_2 = 2$, $p_2 = 16$ and $r_3 = 3$, $p_3 = 24$ to cover a variety of scales. All three parametrization have been applied to the intensity of the input image and are subsequently concatenated, to have a three channel input to the CNN.
%
%
\paragraph{LBP Confidence Calculation:}
The LBP calculation becomes unstable in homogeneous regions because the noise can switch the sign of the differences. Thus, we calculate the confidence, based on the distance to zero-pixel-difference, i.e. the stronger the difference, the higher is the confidence (that the binary number is correct).
This is done by testing against the null hypothesis:
\begin{equation}
	\begin{aligned}
		H_{0_{\textit{LBP}}}\text{: } & \text{The pixel is in a homogeneuos region} \\&\text{i.e } (I_{n_{0}}-I,I_{n_{1}}-I,...,I_{n_{N}}-I)^T = (0,0,...,0)^T.
	\end{aligned}
	\label{eq:H0LBP}
\end{equation}
And homogeneous pixels are assumed to be normal distributed:
\begin{equation}
	\begin{aligned}
		H_{0_{\textit{LBP}}}:\left(\begin{array}{c}
			\hat{I}_{n_{0}}-\hat{I} \\
			\hat{I}_{n_{1}}-\hat{I} \\
			...\\
			\hat{I}_{n_{N}}-\hat{I} \\
		\end{array}\right) \sim N\left(\left(\begin{array}{c}
			0 \\
			0 \\
			...\\
			0 \\
		\end{array}\right),\mathbf{\Sigma}_{\textit{LBP}} \right).
	\end{aligned}
\end{equation}
\paragraph{LBP-Noise Propagation:}
The noise is propagated for the differences $z_i=I_{n_{i}}-I$,  on which the binary numbers $B_i$' are based. By calculating the variance and covariance, the following distribution is obtained (detailed variance and covariance derivation can be found in the Appendix~\ref{app:lbp-prop}):
\begin{align}
	\mathbf{\Sigma}_{LBP} &= \left(\begin{array}{cccc}
		2\sigma^2&	\sigma^2&...&\sigma^2\\
		\sigma^2&	2\sigma^2&...&\sigma^2\\
		...&...&...&...\\
		\sigma^2&\sigma^2&...&2\sigma^2\\
	\end{array}\right).
\end{align}
As the variance and covariance is just depending on the noise deviation, which is the same over the whole image (see Section \ref{sec:noise}) the covariance matrix $\Sigma_{LBP}$ is the same for all pixels in one image.
\begin{figure}[hbtp]
	\begin{center}
		\begin{tabular}{cc}
			&\includegraphics[width=0.48\linewidth]{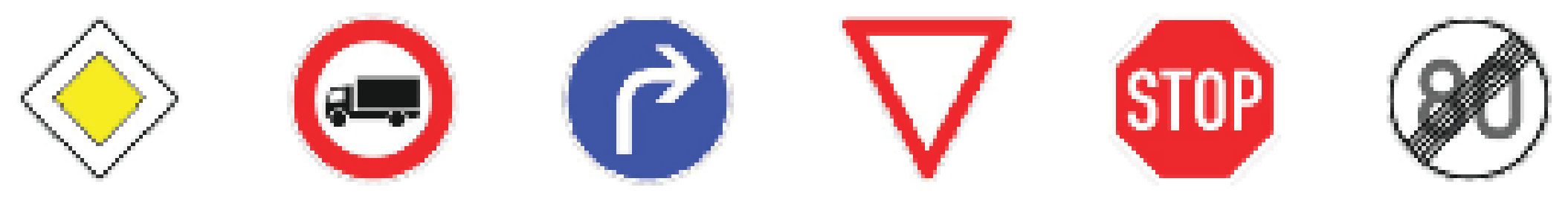}\\
			$P(H_{0_{rg}})$:& \includegraphics[width=0.48\linewidth]{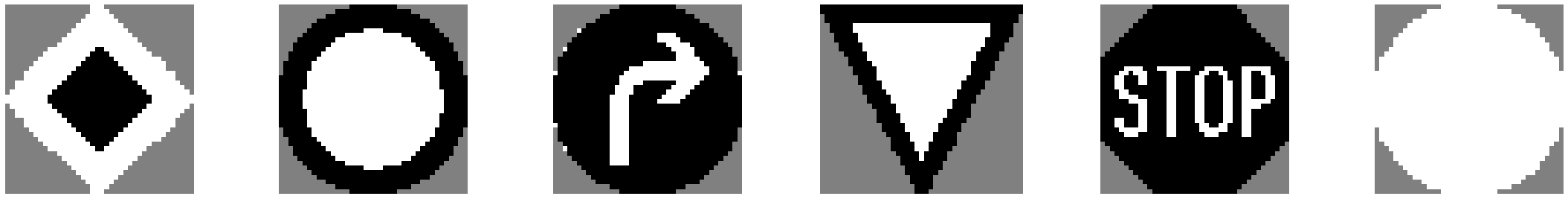}\\
			$P(H_{0_{\textit{LBP}}})$:&\includegraphics[width=0.48\linewidth]{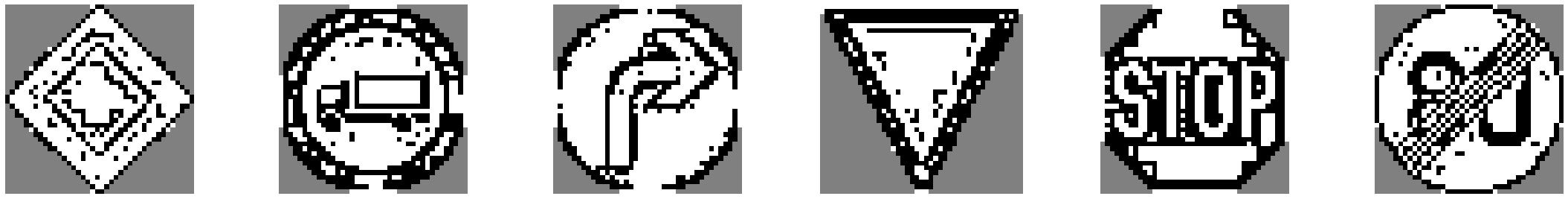}
		\end{tabular}
	\end{center}
	\caption{
    Visualization of $H_0$ for a subset of signs: White pixels indicate regions classified under the null hypothesis (non-colored, homogeneous), while black pixels fall outside it. In the boundary area of the sign, half of the pixels are approximated as belonging to the null hypothesis.} 
	\label{fig:priors}
\end{figure}
\subsection{Dataset Specific Priors}
\label{section:prior}
We estimated the ratio of pixels of the null hypothesis in the GTSRB dataset by analyzing visual unaltered depictions of the signs. For each pixel, we checked if it belongs to the null hypothesis and the ratio relative to the total number of pixels is calculated, as shown in Figure \ref{fig:priors} for some signs.
\section{Experiments} 
\begin{table*}[ht]
	\centering
    \begin{tabular}{lllll}
    	\toprule
    	&\multicolumn{4}{c}{Accuracies on GTSRB dataset}\\
    	&Quasi-invariant operator &supervised & generative & both \\
    	\midrule
    	\midrule
    	Baseline (CNN/VAE) & None &       \textbf{98.62}$\pm$0.41 &         66.72$\pm$4.44 &   \textbf{96.68}$\pm$0.57 \\
    	\midrule
    	Individual Operators 
    	&rg &       90.14$\pm$3.67 &          68.3$\pm$1.56 &   86.74$\pm$4.09 \\
    	&rgConf &        69.0$\pm$1.74 &        27.82$\pm$30.05 &   86.02$\pm$0.39 \vspace*{1pt}\\
    	\cline{2-5}\vspace*{-8pt}\\
    	&LBP &        98.0$\pm$0.19 &        \textbf{86.76}$\pm$0.92 &   \textbf{96.62}$\pm$0.24 \\
    	&LBPConf &       94.92$\pm$0.11 &         84.46$\pm$1.08 &   95.32$\pm$0.28 \\
    	\midrule
    	Decomposed Hybrid Systems
    	&rg+LBP & 98.16$\pm$0.26 &         \textbf{86.38}$\pm$1.36 &   \textbf{96.72}$\pm$0.19 \\
    	&rg+LBPConf &       95.82$\pm$0.38 &         84.08$\pm$2.09 &   95.76$\pm$0.11 \\
    	&rgConf+LBP & 98.12$\pm$0.18 &         \textbf{86.34}$\pm$0.73 &   \textbf{96.64}$\pm$0.21 \\
    	&rgConf+LBPConf &       94.44$\pm$0.59 &          84.3$\pm$2.05 &   95.68$\pm$0.59 \\
    	\bottomrule\\
    \end{tabular}
    
    \caption{
    Results on the GTSRB dataset: accuracies of different operator(rg or LBP) combinations  with and without confidence propagation(Conf). 
    Mean accuracies are \textbf{bolded} if they are the highest or within the 95\% confidence interval of the highest.}
    \label{GTSRB-classification}
\end{table*}
\begin{table*}[ht]
	\centering
	\begin{tabular}{lllll}
		\toprule
		&\multicolumn{4}{c}{Accuracys on \textbf{clustered} GTSRB dataset}\\
		&Quasi-invariant operator & supervised & generative & both \\
		\midrule
		\midrule
		Baseline (CNN/VAE) & None &    \textbf{99.6}$\pm$0.07 &     91.44$\pm$1.84 &  98.88$\pm$0.28 \\
		\midrule
		Individual Operators 
		& rg &    98.9$\pm$0.12 &     97.32$\pm$0.38 &  98.54$\pm$0.25 \\
		& rgConf &    97.92$\pm$0.49 &    54.92$\pm$22.41 &  98.52$\pm$0.23 \vspace*{1pt}\\
		\cline{2-5}\vspace*{-8pt}\\
		& LBP &    99.56$\pm$0.15 &     97.76$\pm$0.54 &  98.82$\pm$0.42 \\
		& LBPConf &    98.46$\pm$0.31 &     97.24$\pm$0.38 &  98.56$\pm$0.18 \\
		\midrule
		Decomposed Hybrid Systems
		&rg+LBP &    \textbf{99.52}$\pm$0.08 &     \textbf{98.58}$\pm$0.15 &  \textbf{99.28}$\pm$0.08 \\
		&rg+LBPConf &    99.36$\pm$0.44 &     97.64$\pm$0.85 &  99.0$\pm$0.12 \\
		&rgConf+LBP &   \textbf{99.62}$\pm$0.11 &     98.2$\pm$0.22 &  \textbf{99.32}$\pm$0.19 \\
		&rgConf+LBPConf &    99.16$\pm$0.23 &     97.26$\pm$0.15 &  98.98$\pm$0.16 \\
		\bottomrule\\
	\end{tabular}
	
	\caption{Results on the clustered-GTSRB dataset, where classes with the same color and shape are merged, showing the accuracies in the same structure as the full GTSRB dataset.}
	\label{GTSRB-classification-small}
\end{table*}

\label{sec:results}
Our empirical investigation follows these key questions:\\
\textbf{Q1) What is the impact of individual quasi-invariant operators with the confidence propagation on classification performance?} We assess the performance of individual quasi-invariant operators (i.e., LBP, \textit{rg}) compared to the baseline CNNs.\\
\textbf{Q2) What is the performance of the decomposed hybrid pipeline?}
We explore whether the combination of multiple operators, leads to improved performance or if accuracy is primarily dependent on the strongest individual operator.\\
\textbf{Q3) Does incorporating quasi-invariant operators improve generalization in low-data regimes?} Data-driven models benefit from extensive labeled samples to learn useful features. We hypothesize that by introducing quasi-invariant operators less data is needed to achieve competitive performance. We test this hypothesis by limiting training data to subsets  where the maximum number of samples per class in the trainset is limited to 5, 10, 50, and 100.
%
%

Before proceeding with the results and analysis, we describe the experimental setup including the training and evaluation approaches, the baselines and the dataset properties. 
For all experiments, the accuracy of the model on the test set was calculated using the model weights that achieved the highest accuracy on the validation set. All training sessions were repeated 5 times, and the mean and standard deviation of the results are reported. Code is available at: \url{github.com/SinaDitzel/DecomposedHybridNetworks}.

\subsection{Experimental Setup}

We evaluate three encoder training configurations: supervised, generative, and combination of both. In the supervised setting, the model is trained end-to-end for classification, with a classifier processing concatenated encoder outputs. In the generative setting, a VAE is trained on the dataset in a first step, and the resulting latent representations are used to train a linear classifier with a frozen VAE encoders. We also evaluate a joint optimization of the classifier and traffic sign reconstruction task. 
We evaluate configurations with and without confidence propagation. When confidence measurements are available, the first two convolutional layers are replaced with normalized convolutional layers. Further details on the neural architectures and hyperparameter settings are provided in Appendix~\ref{app:neural}.

%
The dataset was split into training and test sets using the predefined GTSRB splits \citep{GTSRB}. A validation set was created by separating 15\% of the training samples, with non-random sampling to avoid overlap, ensuring at least 10\% of each class came from complete tracks.
\subsection{Results and Discussion}
\begin{figure*}[ht]
	\begin{minipage}{\columnwidth}
		\centering
		\textbf{GTSRB}
		\vspace*{-3pt}
		\includegraphics[width=\linewidth]{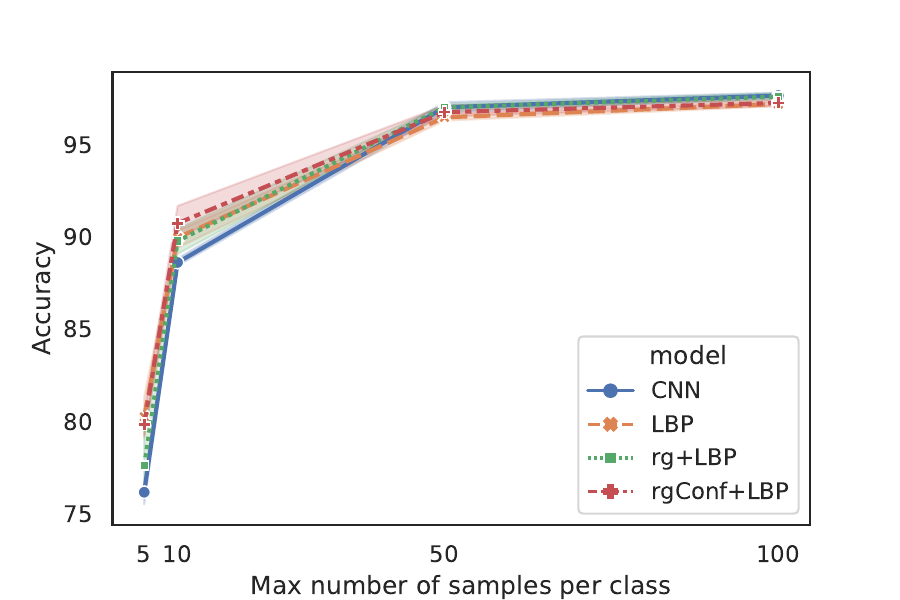}
		
	\end{minipage}
	\begin{minipage}{\columnwidth}
		\centering
		\textbf{clustered-GTSRB}
		\vspace*{-3pt}
		\includegraphics[width=\linewidth]{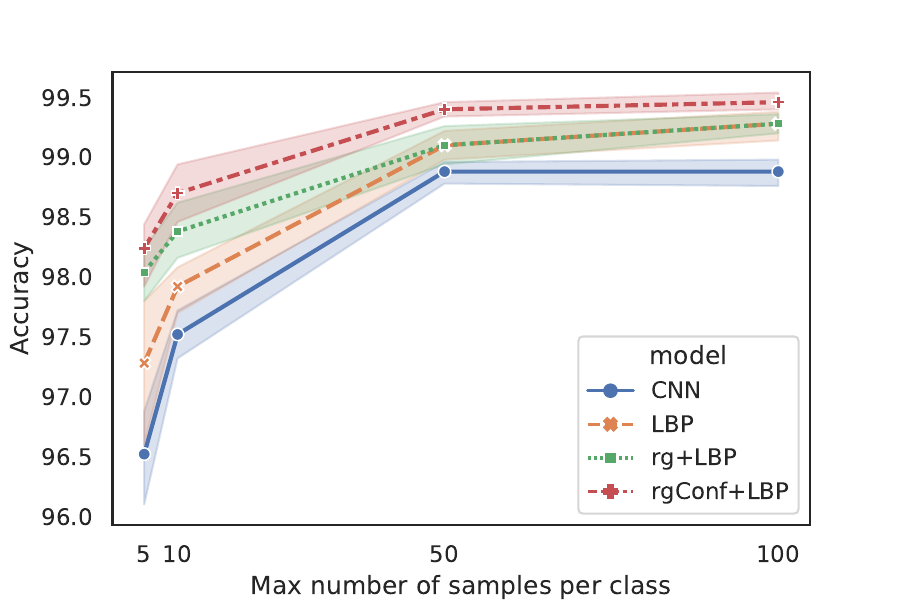}
		
	\end{minipage}
	
	\ \\
	\caption{Classification results in a supervised setting in the limited sample setting on the GTSRB and clustered-GTSRB dataset.}
	\label{GTSRB-small-classification-reduced-samples}
\end{figure*}
\textbf{(Q1) LBP is most suitable operator for the trafic sign domain:}
Table~\ref{GTSRB-classification} presents classification accuracies across different settings. 
In the supervised setting, the CNN has the highest mean accuracy, while the decomposed model demonstrates competitive performance, maintaining accuracy close to the baseline CNN with lower variance across runs.
Among the operators, LBP consistently performed well across all configurations, often matching or surpassing the baseline, particularly in generative training.
In contrast, the \textit{rg} transform demonstrated lower performance, particularly in confidence propagation. This can be attributed to its reliance on color information, which is not always a distinctive feature in GTSRB, where many classes share similar color patterns. These findings suggest that color information alone is insufficient for robust street sign classification. Specifically, confidence propagation assigns low confidence to discriminative, non-color-based features in the image, leading to performance degradation.  To further investigate this, we evaluated the system on a clustered dataset, where the 43 original classes were grouped into 7 clusters based on color and shape. As shown in Table~\ref{GTSRB-classification-small}, the \textit{rg} transform performed significantly better in this setting, confirming that our \textit{rg}-encoder effectively captures color-based features when they are more discriminative. These results highlight the importance of leveraging complementary modalities to achieve competitive performance across diverse domains while also obtaining more disentangled representations, thereby enhancing transparency. 

\textbf{(Q2) A decomposed multi-stream design improves the performance:} Our system integrates multiple complementary features by combining operators that focus on specific modalities. The results indicate that combining parallel encoders with different modalities consistently improves performance. While LBP performs well in supervised and semi-supervised settings, its performance further improves when combined with the \textit{rg} operator. 
A decomposed hybrid design combining both \textit{rg} and LBP achieves the best overall performance in most cases. However, in the supervised case on the non-clustered GTSRB dataset, the CNN outperforms it, highlighting the effectiveness of fully data-driven learning on large dataset.
Among the decomposed hybrid systems, we see that the combination of \textit{rg}, confidence propagation, and LBP yields high accuracies. Particularly on the clustered dataset were this combination has the highest mean accuracy in the supervised and semi-supervised setting.
However, when both operators were combined with confidence propagation, performance slightly declined, likely due to excessive suppression of discriminative information. This highlights the need to balance invariance induction and feature retention to maximize system effectiveness.

\textbf{(Q3) The decomposed design outperforms the baseline in low-sample-setting}
Figure~\ref{GTSRB-small-classification-reduced-samples} visualizes the most effective operator combinations from previous experiments across different sample sizes. Corresponding tables are provided in Appendix~\ref{app:res-limited-samples}. Our decomposed hybrid network \textit{rgConf+LBP} consistently outperforms the baseline CNN, demonstrating superior performance retention in low-data settings due to its introduced invariance. 

While traditional image processing operators such as LBP have proven effective, they are seldom integrated into modern deep learning frameworks. Our results indicate that incorporating such model-based operators can significantly enhance performance. The proposed confidence measure is most effective when the selected modality provides clear class distinctions. 
Furthermore, our findings highlight that the performance of the decomposed hybrid system improves with stronger operator selection. When appropriately chosen, these operators enable the model to surpass the baseline CNN, particularly in data-limited scenarios, underscoring the importance of structured feature extraction in deep learning architectures.
\section{Conclusion}\label{sec:conclusion}


In this work, we explored decomposed hybrid encoder designs, combining quasi invariant operators with neural networks. We introduced a novel confidence measurement leveraging noise estimation, covariance propagation, and hypothesis testing, allowing networks to prioritize features with high confidence during encoding.  Evaluations on the GTSRB  demonstrated the impact of task-appropriate quasi-invariant operators and their complementary effects. Notably, LBP proved to be a strong inductive bias for traffic sign recognition, and a hybrid network integrating the \textit{rg} operator, confidence measurement, and LBP achieved competitive performance, outperforming the baseline CNN in the unsupervised and the low-data settings. These findings highlight the advantages of integrating well understood operators along with their uncertainty estimates into data-driven models. 
Future work should extend this framework to new domains by incorporating additional quasi-invariant operators and explore full uncertainty propagation across network representations and final classifications.


\begin{acknowledgements} 
%

The authors acknowledge the support of various sponsors that made this research possible. This work was supported by the German Federal Ministry of Education and Research (BMBF) funded projects 01IS19062 ”AISEL” .\\ Further for providing time to collaboratively write and finalize the work, Achref Jaziri was supported by the project 45KI16E051 "KIBA"(Künstliche Intelligenz und diskrete Beladeopimierungsmodelle zur Auslastungssteigerung im kombinierten Verkehr ), Iuliia Pluiisch was supported by 16DHBKI019 "ALI",  and S. Ditzel acknowledges the support of Flanders Make (\url{https://www.flandersmake.be}), the strategic research center for the manufacturing industry, and the Flanders AI Research Program (\url{https://www.flandersairesearch.be}).


\end{acknowledgements}

\bibliography{hybrid_network}

\newpage

\onecolumn

\title{Uncertainty-Aware Decomposed Hybrid Networks\\(Supplementary Material)}
\maketitle

\appendix

\section{\texorpdfstring{Derivation of the approximation of the mixture of $\chi^2$ distributions}{Derivation of the approximation of the mixture of chi-squared distributions}}\label{sec:deviation_moments}

As we explained in the main body, for ease of the computation we approximate the uniform mixture of $\chi^2$ distributions by its first two moments. In this section we offer the derivation of these moments.

To reinstate, we defined the pdf $f_{a,b}(x)$ of a uniform mixture of noncentral $\chi^2$ distributions with pdf $f_\chi(x,k,\lambda)$ over a certain range [a,b] as
\begin{align}
	f_{a,b}(x) = \int_{a}^{b} \frac{1}{b-a} f_\chi(x,k,\lambda)  \text{ d}\lambda.
\end{align}

In the following proof we will use the expected value and variance of a noncentral-$\chi^2$ distributed variable $X_\chi$ with the probability density function $f_\chi(x,k,\lambda)$. The expectation and variance are: 
\begin{align}
	\label{math:exp_noncentral chi square}
	E[X_\chi]=\int_{-\infty}^{\infty}xf_\chi(x,k,\lambda)\text{ d}x = k+\lambda,
\end{align}
\begin{equation}
	\begin{aligned}
		\label{math:var_noncentral chi square}
		Var[X_\chi]= \int_{-\infty}^{\infty}(x-E[X_\chi])^2 f_\chi(x,k,\lambda)\text{ d}x \\
		= 2(k+2\lambda).
	\end{aligned}
\end{equation}
For the uniform mixture of noncentral-$\chi^2$ distribution, we will first show the derivation of the expectation. For a random variable $X$,  with pdf $f_{a,b}(x)$ the expectation E[X] is: 
\allowdisplaybreaks
\begin{align}
	E[X]\stackrel{\ref{math:exp_cont}}{=}
	&\int_{-\infty}^{\infty}xf_{a,b}(x)\text{ d}x\\
	=&\int_{-\infty}^{\infty}x \int_a^b \frac{1}{b-a} f_\chi(x,k,\lambda)  \text{ d}\lambda \text{ d}x \notag\\
	=&\frac{1}{b-a} \int_a^b \int_{-\infty}^{\infty}xf_\chi(x,k,\lambda)  \text{d}x \text{d}\lambda\notag\\
	\stackrel{\ref{math:exp_noncentral chi square}}{=}&\frac{1}{b-a} \int_a^b (k+\lambda) \text{ d}\lambda  \notag\\
	=& \frac{1}{b-a}\left[k\lambda+\frac{\lambda^2}{2} \right]_{\lambda= a}^{\lambda= b}\notag\\
	=& \frac{(kb+\frac{b^2}{2}-ka-\frac{a^2}{2})}{b-a} = k+\frac{a+b}{2}.
\end{align}
Next, we will calculate the variance of $X$. Therefore, we show first that the variance can be transformed to three terms, by expanding the squared parenthesis and multiply all sub-terms with $f_{a,b}(x)$: 
\begin{equation}
	\begin{aligned}
		Var[X]
		\stackrel{\ref{math:var_cont}}{=}& \int_{-\infty}^{\infty}(x-E[X])^2 f_{a,b}(x)\text{d}x\\
		=& \int_{-\infty}^{\infty}(x^2-2xE[X]+E[X]^2)f_{a,b}(x)\text{d}x\\
		=& \int_{-\infty}^{\infty}x^2 f_{a,b}(x)-2xE[X]f_{a,b}(x)+E[X]^2f_{a,b}(x)\text{d}x\\
		=& \underbrace{\int_{-\infty}^{\infty}x^2 f_{a,b}(x) \text{d}x}_{\text{see \ref{eq:var_1}}} -\underbrace{\int_{-\infty}^{\infty}2xE[X]f_{a,b}(x)\text{d}x}_{\text{see \ref{eq:var_2}}}\\
		&+ \underbrace{\int_{-\infty}^{\infty}E[X]^2f_{a,b}(x)\text{d}x}_{\text{see \ref{eq:var_3}}}. \\
	\end{aligned}
\end{equation}
The first term can be expand to three terms, by adding $0= -2x(k+\lambda)+(k+\lambda)^2 +2x(k+\lambda)-(k+\lambda)^2$ to $x^2$:

\begin{equation}
	\begin{aligned}
		x^2\label{math:x-extension} &= x^2- 2x(k+\lambda)+(k+\lambda)^2 +2x(k+\lambda)-(k+\lambda)^2 \\ 
		&=((x-(k+\lambda))^2 +2x(k+\lambda)-(k+\lambda)^2).&
	\end{aligned}
\end{equation}

This allows us to rewrite the formula to contain the formulas for $E[X_\chi]$ and $\textit{Var}[X_\chi]$ and an integral over a pdf that sums to 1, to simplify the formula, that we can then integrate.

\begin{align}
	\int_{-\infty}^{\infty}x^2 f_{a,b}(x) \text{d}x\label{eq:var_1} =&\int_{-\infty}^{\infty}x^2 \int_a^b \frac{1}{b-a}f_\chi(x,k,\lambda)  \text{ d}\lambda \text{d}x\nonumber\\
	=&\frac{1}{b-a} \int_a^b \int_{-\infty}^{\infty}x^2  f_\chi(x,k,\lambda)  \text{d}x\text{d}\lambda\nonumber\\
	\stackrel{\ref{math:x-extension}}{=}&\frac{1}{b-a} \int_a^b \underbrace{\int_{-\infty}^{\infty}(x-(k+\lambda))^2   f_\chi(x,k,\lambda)  \text{d}x}_{\stackrel{\ref{math:var_noncentral chi square}}{=}2(k+2\lambda)}\text{d}\lambda \nonumber+ \frac{2}{b-a} \int_a^b (k+\lambda)   \underbrace{\int_{-\infty}^{\infty} xf_\chi(x,k,\lambda)  \text{d}x}_{\stackrel{\ref{math:exp_noncentral chi square}}{=}k+\lambda}\text{d}\lambda\nonumber\\
	&-\frac{1}{b-a} \int_a^b  (k+\lambda)^2 \underbrace{\int_{-\infty}^{\infty}  f_\chi(x,k,\lambda)  \text{d}x}_{=1}\text{d}\lambda\nonumber\\
	=&\frac{1}{b-a} \int_a^b 2(k+2\lambda) \text{ d}\lambda + \frac{2}{b-a} \int_a^b (k+\lambda)^2\text{d}\lambda\nonumber-\frac{1}{b-a} \int_a^b (k+\lambda)^2  \text{d}\lambda\nonumber\\
	=&\frac{2}{b-a} \int_a^b k+2\lambda \text{ d}\lambda  +\frac{1}{b-a} \int_a^b (k+\lambda)^2  \text{d}\lambda\nonumber\\
	=&\frac{2}{b-a}\left[k\lambda+\lambda^2\right]_{\lambda=a}^{\lambda=b}\nonumber + \frac{1}{b-a} \left[k^2\lambda+k\lambda^2+1/3\lambda^3\right]_{\lambda=a}^{\lambda=b}\nonumber\\
	=&\frac{2kb-2ka+2b^2-2a^2}{b-a}\nonumber+\frac{k^2b-k^2a+kb^2-ka^2+\frac{1}{3}b^3-\frac{1}{3}a^3}{b-a}\nonumber\\
	=&\frac{(2k+k^2)b-(2k+k^2)a+(2+k)b^2-(2+k)a^2}{b-a}\nonumber+\frac{\frac{1}{3}b^3-\frac{1}{3}a^3}{b-a}\nonumber\\
	=&\frac{(2k+k^2)(b-a)}{b-a}+\frac{(2+k)(b^2-a^2)}{b-a}\nonumber\frac{(\frac{1}{3})(b^3-a^3)}{b-a}\nonumber\\
	=&(2k+k^2)+(2+k)(a+b)+ \frac{1}{3}(a^2+ab+b^2).
\end{align}
The second term we can move the constant outside the integral, so that only the expectation $E[X]$  remains in the integral:

\begin{equation}
	\begin{aligned}
		\int_{-\infty}^{\infty}2xE[X]f_{a,b}(x)\text{d}x\label{eq:var_2}&= \int_{-\infty}^{\infty}2xE[X]\int_a^b \frac{1}{b-a}f_\chi(x,k,\lambda)  \text{ d}\lambda\text{d}x \\
		&=2E[X]\int_{-\infty}^{\infty}x\int_a^b  \frac{1}{b-a} f_\chi(x,k,\lambda)  \text{d}\lambda\text{d}x \\
		&=2E[X]E[X]\\ &=2E[X]^2.
	\end{aligned}
\end{equation}

In the third term, we can use again that a integral over all possible outcomes of a pdf sums up to 1, such that 

\begin{equation}
	\begin{aligned}
		\int_{-\infty}^{\infty}E[X]^2f_{a,b}(x)\text{d}x
		=E[X]^2\int_{-\infty}^{\infty}f_{a,b}(x)\text{d}x
		=E[X]^2.\label{eq:var_3}
	\end{aligned}
\end{equation}

Combining the transformations, we then gain the short equation for the variance, presented in the main paper:
\begin{equation}
	\begin{aligned}
		\textit{Var}[X]=& \underbrace{\int_{-\infty}^{\infty}x^2 f_{a,b}(x) \text{d}x}_{\text{see \ref{eq:var_1}}} -\underbrace{\int_{-\infty}^{\infty}2xE[X]f_{a,b}(x)\text{d}x}_{\text{see \ref{eq:var_2}}}+ \underbrace{\int_{-\infty}^{\infty}E[X]^2f_{a,b}(x)\text{d}x}_{\text{see \ref{eq:var_3}}} \\
		=&(2k+k^2)+(2+k)(a+b) +\frac{1}{3}(a^2+ab+b^2)-2E[X]^2 +E[X]^2\\
		=&2k+k^2+(2+k)(a+b)+\frac{1}{3}(a^2+ab+b^2)-\left(k+\frac{a+b}{2}\right)^2\\
		=&2k+2(a+b)+\frac{1}{3}(a^2+ab+b^2)-\frac{1}{4}\left(a^2+2ab+b^2\right)\\
		=&2k+2(a+b)+\frac{1}{12}(a^2+2ab+b^2).
	\end{aligned}
\end{equation}

\section{Variance and Covariance of LBP distribution} \label{app:lbp-prop}
In the main paper, we chose LBP as one of the quasi-invariances to use. For its computation, intensity $I$ is first computed from the RGB value, from which distance $z_i = I_{n_{i}}-I$ is computed to the  neighboring pixel intensities $I_{n_{i}}$. In order to compute the confidence that these distances $\mathbf{z}$ are relevant, we chose zero-centered Gaussian noise as our intensity noise model: $\hat{I} = I+ \eta$ with $\eta \sim \mathcal{N}(0,\sigma^2)$.

Hence, the random variable $\hat{Z}_{i}$ (from which we observe $\hat{z}_i$) can be expressed as:

\begin{align}
	\hat{Z}_{i} = \hat{I}_{n_{i}}-\hat{I} = (I_{n_{i}}+\eta_i)-(I+\eta).
\end{align}

In the main paper we stated the covariance matrix for these distances, which we will derive here. As a covariance matrix is a square matrix that encapsulates the variances and covariances, we will derive $\textit{Var}[\hat{Z}_{i}]$ and $\textit{Cov}[\hat{Z}_{i}, \hat{Z}_{j}]$ for pixel $\hat{Z}_{i}$ and neighbbour $\hat{Z}_{j}$.

\subsection{Definitions and Formulas for Variance and Expectation}

First, we provide definitions and formulas for variance and expectation, which are basic concepts in probability theory and statistics, and which we will use in the following sections.\\

The variance of a random variable $X$ is a measure of how much the values of $X$ vary from its expected value. The variance is denoted by Var[X] and is defined as
\begin{align}
	\textit{Var}[X] = E[(X - E[X])^2].
\end{align}	
where $E[X]$ is the expected value of $X$.\\

The covariance between two random variables $X$ and $Y$ is a measure of how much the two variables vary together. The covariance is denoted by $\textit{Cov}[X,Y]$ and is defined as
\begin{align}
	\textit{Cov}[X,Y] = E[(X - E[X])(Y - E[Y])].
\end{align}
\ \\
For a continuous variable $X$ with pdf $f(x)$ the expectation is defined as 
\begin{align}
	\label{math:exp_cont}
	E[X] = \int_{-\infty}^{\infty}xf(x)\text{ d}x.
\end{align} 

and the variance can be calculated with
\begin{align}
	\label{math:var_cont}
	\textit{Var}[X] = \int_{-\infty}^{\infty}(x-E[X])^2 f(x)\text{ d}x.
\end{align}

\subsection{Variance}
\begin{equation}
	\begin{aligned}
		\textit{Var}[\hat{Z}_{i}] &= \textit{Var}[(\hat{I}_{n_{i}}-\hat{I})^{2}]\\
		&= \textit{Var}[\hat{I}_{n_{i}}]+\textit{Var}[\hat{I}]-2Cov[\hat{I}_{n_{i}},\hat{I}]\\
		&= \textit{Var}[\hat{I}_{n_{i}}]+\textit{Var}[\hat{I}]\\
		&= 2\sigma^2.
	\end{aligned}
\end{equation}

\subsection{Covariance}
Next we compute the covariance between two distances (of the center pixel to two different  neighbors).
We will use that the difference between the variable of the distance  $\hat{Z}_{i}$ to its expected value $z_i$ can be expressed as the difference between the noise terms:
\begin{equation}
	\begin{aligned}
		\hat{Z}_{i}-z_{i} &= (I_{n_{i}}+\eta_i-(I+\eta)) -(I_{n_{i}} -I)\\ &=\eta_i -\eta.
	\end{aligned}
\end{equation}

We calculate the covariance as
\begin{equation}
	\begin{aligned}
		\textit{Cov}[\hat{Z}_{i}, \hat{Z}_{j}] &=E[(\hat{Z}_{i}-z_{i})(\hat{Z}_{j}-z_{j})]\\
		&=E[(\eta_i -\eta )(\eta_j -\eta)]\\
		&=E[\eta_i\eta_j] + E[\eta_i(-\eta)] + E[\eta_j (-\eta)] + E[\eta^2]\\
		&= 0+0+0 + \sigma^2 = \sigma^2.
	\end{aligned}
\end{equation}

This holds due to the relationship between the noise factors: $E[\eta_i\eta_j] = E[\eta_i\eta]= E[\eta_j\eta]= 0$, as $\eta_i\eta_j\eta$ are independent:

\begin{equation}
	\begin{aligned}
		E[\eta_i\eta_j] &=E[((I_{n_{i}}+\eta_i) - I_{n_{i}})((I_{n_{j}}+\eta_j)-I_{n_{j}})]\\
		&=\textit{Cov}[\hat{I}_{n_{i}},\hat{I}_{n_{j}}]=0,\\
	\end{aligned}
\end{equation}
and $E[\eta\eta]=\sigma^2$:

\begin{equation}
	\begin{aligned}
		E[\eta\eta] &=E[((I+\eta)-I)^2]\\
		&=\textit{Var}[\hat{I}]=\sigma^2.\\
	\end{aligned}
\end{equation}

\section{Further Implementation details} \label{app:implementation}
In this section, we provide additional information on the implementation of our method, including details on the noise estimation, illumination correction, setting appropriate $\lambda$ values, and calculating priors. Further, we detail the architecture and hyperparameters of the neural network.
\subsection{Noise estimation} \label{app:noise}
As mentioned in the main body we assume the noise distribution can be approximated by deviations in homogeneous region. We implemented a noise estimation method\citep{noise_immerkaer_extended}, which first uses the Sobel operator for edge detection to detect areas with less structure, then a Laplacian operator is additionally applied which suppresses image structures, so that the residuals can be used to calculate the standard deviation.
We used the 10\% of the area with the least structure. Additionally, 5\% of pixels with the highest and lowest intensity and saturation have also been discarded for more stability. A Canny-edge detector was used with thresholds $t_1 = 55\%$ and $t_2 = 75\%$, the neighborhood size $n_s$ has been set to 5 pixels.

\subsection{Illumination Correction} \label{app:color_correction}
Before we use the \textit{rg}-transform we first apply a illumination correction, to achieve (more) neutral illumination.
Based on the von Kries Model\citep{kries_1970_illumination_correction} 
the image $(R_{\mathbf{e}^\prime},G_{\mathbf{e}^\prime},B_{\mathbf{e}^\prime})^T$ taken under an unknown light illumination $\mathbf{e}^\prime = (e^\prime_R,e^\prime_G, e^\prime_B)^T$ can be transformed to image $(R_\mathbf{e}, G_\mathbf{e}, B_\mathbf{e})^T$ taken under the canonical illuminant $\mathbf{e}$ by linear transformation, that independently scales the three recorded color channels:
\begin{equation}
	\begin{aligned}
		\left(\begin{array}{c}
			R_\mathbf{e}\\G_\mathbf{e}\\B_\mathbf{e}\\
		\end{array}\right)=\left(\begin{array}{c c c}
			\frac{e_R}{e^\prime_R}&0&0\\
			0&\frac{e_G}{e^\prime_G}&0\\
			0&0&\frac{e_B}{e^\prime_B}\\
		\end{array}\right) \left(\begin{array}{c}
			R_{\mathbf{e}^\prime}\\G_{\mathbf{e}^\prime}\\B_{\mathbf{e}^\prime}\\
		\end{array}\right).
	\end{aligned}
\end{equation}

The illumination $\mathbf{e}^\prime$ is estimated based on the assumption, that an image consists of numerous gray ($R=G=B$) pixels. Especially for street signs this assumption holds because every sign consists of white areas. In the image taken under illumination $e^\prime$ these gray values might be slightly shifted. The deviation, introduced by the color temperature of the illuminant, is used to approximate $e^\prime$, similar to \citep{huo_2006_white_balance_gray_color_points}. In the normalized rg space all gray values ($R=G=B$) are given by $\textbf{rg} = (\frac{1}{3}, \frac{1}{3})^T$. Thus, for each color channel the mean $\bar{r}$ and $\bar{g}$ of values in a specified range $\delta$ around the gray point $[\frac{1}{3}-\delta, \frac{1}{3}+\delta]$ is calculated, and assumed to be the light color in \textit{rg} space. Leading to the following illumination correction:
\begin{equation}
	\begin{aligned}
		\left(\begin{array}{c}
			r\\g\\
		\end{array}\right)=\left(\begin{array}{c c c}
			\frac{1/3}{\bar{r}}&0&0\\
			0&\frac{1/3}{\bar{g}}&0\\
		\end{array}\right) \left(\begin{array}{c}
			r^\prime\\g^\prime\\
		\end{array}\right).
	\end{aligned}
\end{equation}

\subsection{\texorpdfstring{Setting $\lambda$ values}{Setting lambda values}}\label{app:lambda}
In the main body, we explained that we approximate $H_0$ and $\neg H_0$ by a uniform mixture of noncentral-$\chi^2$-distribution with noncentrality parameters $\lambda$ ranging from $\lambda_1$ to $\lambda_2$. For $H_0$ we set $\lambda$ to range from $\lambda_{2_{\neg H0}}=0$ (representing small deviations from centrality) to $\lambda_{2_{H0}}$. For $\neg H_0$, the noncentrality parameter starts from $\lambda_{2_H0}$, such that $\lambda_{2_{H0}} = \lambda_{1_{\neg H0}}$, indicating a clear departure from centrality.
Selecting appropriate values for $\lambda_{2_{H0}} = \lambda_{1_{\neg H0}}$ and $\lambda_{2_{\neg H0}}$ is challenging, as deviation from the mean depends on the conditions under which the images were captured. To address this, we compute multiple $\lambda$ values per image, including one adaptive value based on the distribution of Mahalanobis distances ($\hat{d}$). \\
Collecting all $\hat{d}$ for an image into the vector $\hat{\textbf{d}}$, we choose $\lambda_{2_{H0}} = \lambda_{1_{\neg H0}}$ values as $0, 100, 1000, \frac{1}{2} \textit{median}(\hat{\textbf{d}})$ and compute a weighted mean from the resulting confidence maps. To capture high deviations $\lambda_{2_{\neg H0}}$ is set to $1.75\cdot \textit{median}(\hat{\textbf{d}})$. 

\subsection{Prior Calculation} \label{app:priors}
We estimated the ratio of pixels belonging to the null hypothesis in the GTSRB dataset. To analyze the ratio, we use visual (unaltered) depictions of the signs instead of recorded images. We check for each pixel of the sign if it belonging to the null hypothesis and calculated the ratio of those to the total number of pixels. We assume all pixels with $\textit{r}=\frac{1}{3}$ and $\textit{g}=\frac{1}{3}$ as not colored and all pixels with no difference to the  neighbors $\textit{sum}(z_0,...z_N)=0$ as homogeneous according to the definitions of null hypothesis Equation~\ref{eq:H0rg} and Equation~\ref{eq:H0LBP}. For the boundary area around the sign, which is unknown, we approximate an equal distribution of pixels belonging to the null hypothesis $P(H_0)$=50\%. This leads to the masks visualized in Figure~\ref{fig:priors} for a subset of the signs. We average those over all signs to calculate the priors, which leads to $P(H_{0_{rg}})= 47.2 \%$ and
$P(H_{0_{\textit{LBP}_{1,8}}})= 54.45\%$, $P(H_{0_{\textit{LBP}_{3,24}}})= 52.2\%, P(H_{0_{\textit{LBP}_{5,40}}})= 50.36\%$ on the GTSRB dataset.

\subsection{Neural Architectures} \label{app:neural}
Each operator is paired with an encoder, and we evaluated three different model configurations: supervised CNN, generative VAE, and combination of both. The supervised setting involves a multilayer perceptron (MLP) that takes the (concatenated) outputs of the encoders as input. In the semi-supervised configuration, we optimize both the MLP and the reconstruction of the traffic signs, leveraging the latent representations learned by the encoders. In the generative setting, a VAE is trained on the dataset in a
first step, and the resulting latent representations are used to
train a linear classifier with a frozen VAE encoders.
We evaluate different combinations with and without confidence propagation. In scenarios where the input data includes confidence measurements, the first two convolutional layers of all architectures are replaced with normalized convolutional layers\citep{NCNN}. Additionally, confidence pooling is applied during subsampling. The CNN architecture is adapted from OneCNN \citep{OneCNN}, their architecture achieved over 99\% on the GTSRB dataset. Our adaptation preserves the same number of layers and kernel sizes but excluding skip connections. We also used a patch size of 48 × 48 and a batch size of 64, consistent with the settings in OneCNN \citep{OneCNN}. The VAE architecture consists of an encoder with 5 convolutional layers and a decoder with 5 transposed convolutional layers. The input size for the VAE is set to 128 × 128 to enable more effective feature reconstruction. Training is performed using the Adam optimizer with a learning rate of $10^{-3}$ for 50 epochs, and the Kaiming weight initialization\citep{kaimingUniform} is applied for improved convergence.
The model architectures are described in Table~\ref{tab:architectures}.
\begin{table}[tbh]
	\centering
	\begin{tabular}{|c|c|c|}
		\hline
		\textbf{Architecture} & \textbf{Layer Type} & \textbf{Details} \\ \hline
		
		\multirow{6}{*}{CNN} 
		& Convolution & 64 channels, 5x5 kernel, stride 1, no padding \\ \cline{2-3} 
		& Max-Pooling & 2x2 window, stride 2 \\ \cline{2-3} 
		& Convolution & 128 channels, 3x3 kernel, stride 1, no padding \\ \cline{2-3} 
		& Max-Pooling & 2x2 window, stride 2 \\ \cline{2-3} 
		& Convolution & 128 channels, 3x3 kernel, stride 1, no padding \\ \cline{2-3} 
		& Max-Pooling & 2x2 window, stride 2 \\ \hline
		
		\multirow{5}{*}{VAE Encoder} 
		& Convolution & 128 channels, 3x3 kernel, stride 2, padding 1 \\ \cline{2-3} 
		& Convolution & 256 channels, 3x3 kernel, stride 2, padding 1 \\ \cline{2-3} 
		& Convolution & 512 channels, 3x3 kernel, stride 2, padding 1 \\ \cline{2-3} 
		& Convolution & 1024 channels, 3x3 kernel, stride 2, padding 1 \\ \cline{2-3} 
		& Latent Space & Outputs \(\mu\) and \(\sigma\) for  \(z\) with dimension 128. \\ \hline
		
		\multirow{5}{*}{VAE Decoder} 
		& Transposed Convolution & 512 channels, 3x3 kernel, stride 2, padding 1 \\ \cline{2-3} 
		& Transposed Convolution & 256 channels, 3x3 kernel, stride 2, padding 1 \\ \cline{2-3} 
		& Transposed Convolution & 128 channels, 3x3 kernel, stride 2, padding 1 \\ \cline{2-3} 
		& Transposed Convolution & 3 channels, 3x3 kernel, stride 2, padding 1 \\ \hline
		
		\multirow{5}{*}{MLP Classifier} 
		& Dropout & Probability 0.25 \\ \cline{2-3} 
		& Linear & Input: classifier\_input\_dim, Output: 128 \\ \cline{2-3} 
		& Linear & Non-linearity \\ \cline{2-3} 
		& Dropout & Probability 0.25 \\ \cline{2-3} 
		& Linear & Input: 128, Output: num\_classes \\ \hline
		
	\end{tabular}
	\caption{Summary of CNN, VAE, and MLP Classifier Architectures. Convolutional layers (and equivalent for transposed convolution layers) for the supervised and semi-supervised training are followed by Batch Normalization and ReLU activation. In the pre-trained VAE only ReLU activation without Batch Normalization is used. In the VAE Encoder, the final layer computes \(\mu\) and \(\sigma\) for the latent variable \(z\) with dimension 128.  In scenarios where the input data includes confidence measurements, the first two convolutional layers of all architectures are replaced with normalized convolutional layers and the first two poolings with confidence pooling}
	\label{tab:architectures}
\end{table}

\section{Further Experimental Results}
\subsection{Limited Sample Setting}\label{app:res-limited-samples}
\begin{figure}[ht]
	\centering
	\begin{tabular}{|>{\centering\arraybackslash}p{0.16\linewidth} 
			|>{\centering\arraybackslash}p{0.04\linewidth} 
			|>{\centering\arraybackslash}p{0.12\linewidth} 
			|>{\centering\arraybackslash}p{0.20\linewidth} 
			|>{\centering\arraybackslash}p{0.15\linewidth} 
			|>{\centering\arraybackslash}p{0.08\linewidth} 
			|>{\centering\arraybackslash}p{0.06\linewidth}|}
		\hline
		{\centering
		 \includegraphics[width=0.99\linewidth, trim=0 0 0 -5]{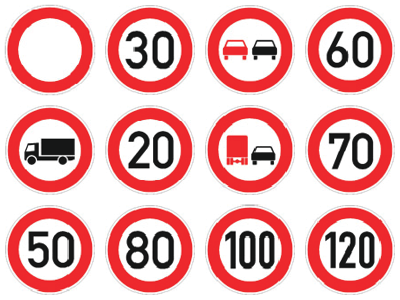}} &
		{\centering \includegraphics[width=0.99\linewidth]{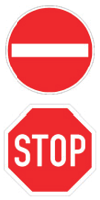}} &
		{\centering \includegraphics[width=0.99\linewidth]{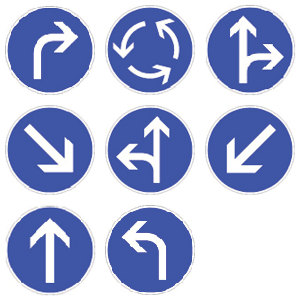}} &
		{\centering \includegraphics[width=0.99\linewidth]{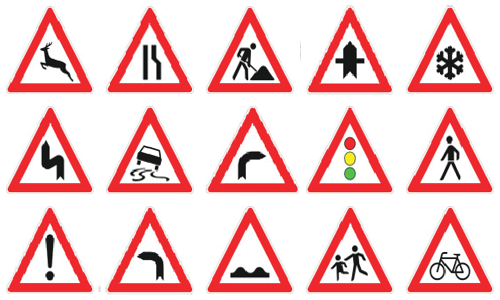}} &
		{\centering \includegraphics[width=0.25\linewidth]{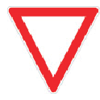}} &
		{\centering \includegraphics[width=0.99\linewidth]{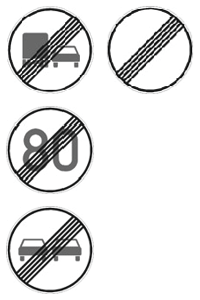}} &
		{\centering \includegraphics[width=0.75\linewidth]{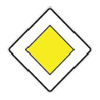}} \\
		
		{\centering Circle with Red Margin} & 
		{\centering Filled Red} & 
		{\centering Blue Filled Circle} & 
		{\centering Triangle with Red Surrounding \\(pointing up) } & 
		{\centering Triangle with Red Surrounding (pointing down)} & 
		{\centering Gray circle} & 
		{\centering Yellow Rectangle} \\
		\hline
	\end{tabular}
	\vspace*{7pt}
	\caption{Classes of the clustered GTSRB dataset}
	\label{tab:cluster_collages}
\end{figure}

In Table~\ref{tab:limited_sample_gtsrb} are some more detailed results on the GTSRB dataset for the limited sample setting described in the main body. Figure~\ref{tab:cluster_collages} visualize  how the dataset is clustered for the experiments on the clustered dataset and Table~\ref{tab:limited_sample_gtsrb_small} shows the detailed result for the clustered dataset. 
\begin{table}[ht]
	\centering
\begin{tabular}{lllll}
	\toprule
	Method & \#samples=5 & \#samples=10 &  \#samples=50 & \#samples=100 \\
	\midrule
	CNN &        76.16$\pm$0.79 &          88.6$\pm$0.25 &          97.0$\pm$0.37 &          97.64$\pm$0.21 \\
	rg &         53.2$\pm$1.73 &         62.96$\pm$1.07 &         84.26$\pm$0.67 &           88.3$\pm$0.38 \\
	rgConf &        42.36$\pm$0.86 &         44.94$\pm$0.94 &         52.66$\pm$0.66 &          55.72$\pm$0.87 \\
	LBP &        80.24$\pm$1.49 &         89.94$\pm$0.61 &         96.46$\pm$0.23 &          97.18$\pm$0.16 \\
	LBPConf &         66.28$\pm$1.9 &         75.42$\pm$0.76 &         86.68$\pm$0.89 &          89.22$\pm$0.46 \\
	rg+LBP &         77.6$\pm$2.12 &          89.76$\pm$0.9 &          97.0$\pm$0.21 &          97.58$\pm$0.23 \\
	rg+LBPConf &         65.5$\pm$2.46 &         76.44$\pm$1.68 &         90.44$\pm$0.78 &           92.9$\pm$0.47 \\
	rgConf+LBP &        79.84$\pm$0.76 &         90.72$\pm$1.05 &         96.74$\pm$0.38 &          97.24$\pm$0.21 \\
	rgConf+LBPConf &        68.82$\pm$1.91 &         76.38$\pm$0.47 &         87.72$\pm$0.94 &          89.86$\pm$0.44 \\
	\bottomrule
\end{tabular}
\caption{Accuracies where the maximum number of samples per class is limited to p = 5,10,50,100 on the GTSRB dataset}
\label{tab:limited_sample_gtsrb}
\end{table}

\begin{table}
	\centering
	\begin{tabular}{lllll}
		\toprule
		Method & \#samples=5 & \#samples=10 &  \#samples=50 & \#samples=100 \\
		\midrule
		CNN &         96.52$\pm$0.5 &         97.52$\pm$0.26 &         98.88$\pm$0.11 &          98.88$\pm$0.13 \\
		rg &         96.02$\pm$0.3 &         96.22$\pm$0.55 &         97.94$\pm$0.23 &          98.28$\pm$0.13 \\
		rgConf &          93.0$\pm$0.6 &         94.56$\pm$0.88 &         96.52$\pm$0.22 &          97.16$\pm$0.21 \\
		LBP &        97.28$\pm$0.75 &         97.92$\pm$0.25 &          99.1$\pm$0.16 &          99.28$\pm$0.16 \\
		LBPConf &        90.84$\pm$0.41 &         93.68$\pm$0.64 &          96.5$\pm$0.23 &          97.12$\pm$0.18 \\
		rg+LBP &         98.04$\pm$0.3 &          98.38$\pm$0.3 &           99.1$\pm$0.2 &          99.28$\pm$0.11 \\
		rg+LBPConf &        97.12$\pm$0.25 &         97.38$\pm$0.38 &          98.5$\pm$0.14 &          99.02$\pm$0.08 \\
		rgConf+LBP &        98.24$\pm$0.36 &          98.7$\pm$0.28 &          99.4$\pm$0.07 &          99.46$\pm$0.09 \\
		rgConf+LBPConf &         96.2$\pm$0.46 &         96.92$\pm$0.43 &         98.46$\pm$0.22 &          98.94$\pm$0.11 \\
		\bottomrule
	\end{tabular}
	\caption{Accuracies where the maximum number of samples per class is limited to p = 5, 10, 50, 100 on the \textbf{clustered GTSRB dataset}}
	\label{tab:limited_sample_gtsrb_small}
\end{table}

\end{document}